\documentclass{article}

\PassOptionsToPackage{numbers}{natbib}
\usepackage{subcaption}
\usepackage{wrapfig}

 \usepackage[preprint]{neurips_2025}

\usepackage{adjustbox}
\usepackage{parskip}
\usepackage[utf8]{inputenc} 
\usepackage[T1]{fontenc}    
\usepackage{hyperref}       
\usepackage{url}            
\usepackage{booktabs}       
\usepackage{amsfonts}       
\usepackage{nicefrac}       
\usepackage{microtype}      
\usepackage{xcolor}         

\usepackage{amsmath}
\usepackage{amssymb}
\usepackage{tikz}
\usepackage{tikz-layers}
\usepackage{xcolor}
\usepackage{graphicx}
\usepgflibrary{patterns}
\usetikzlibrary{arrows.meta,positioning,calc,shapes.geometric,decorations.text,patterns,patterns.meta,fit,backgrounds,chains,shadows,math,circuits.ee.IEC,decorations.pathmorphing, decorations.pathreplacing, decorations.shapes}

\definecolor{limegreen}{HTML}{97c65a}
\definecolor{brickred}{HTML}{b92622}
\definecolor{midnightblue}{HTML}{005c7f}

\newcommand{\rmS}{\mathbf{S}}  
\newcommand{\rmK}{\mathbf{K}}  
\newcommand{\rmV}{\mathbf{V}}  
\newcommand{\rmQ}{\mathbf{Q}}  
\newcommand{\E}{\mathbb{E}}

\newcommand{\LL}{\mathcal{L}}

\newcommand{\rd}{{\mathrm{d}}}

\newcommand{\bv}{{\mathbf{v}}}

\newcommand{\eps}{{\boldsymbol \varepsilon}}

\title{ARFlow: Autoregressive  Flow with Hybrid Linear Attention}

\author{Mude Hui\textsuperscript{\rm 1}\footnotemark[1]
\quad  Rui-Jie Zhu\textsuperscript{\rm 1}\footnotemark[1]
\quad  Songlin Yang\textsuperscript{\rm 2}\footnotemark[1] 
\\
\quad Yu Zhang\textsuperscript{\rm 3}
\quad  Zirui Wang\textsuperscript{\rm 4}
\quad Yuyin Zhou\textsuperscript{\rm 1}
\quad Jason Eshraghian\textsuperscript{\rm 1}
\quad  Cihang Xie\textsuperscript{\rm 1} \vspace{.1em}
  \\\small $^{\star}$equal technical contribution\vspace{.3em} \\
\textsuperscript{\rm 1}{UC Santa Cruz}\qquad
\textsuperscript{\rm 2}{MIT}\qquad 
\textsuperscript{\rm 3}{Bitdeer}\qquad
\textsuperscript{\rm 4}{Apple}\quad \vspace{-.2em}
}

\begin{document}

\maketitle

\begin{abstract}

Flow models are effective at progressively generating realistic images, but they generally struggle to capture long-range dependencies during the generation process as they compress all the information from previous time steps into a single corrupted image.
To address this limitation, we propose integrating autoregressive modeling---known for its excellence in modeling complex, high-dimensional joint probability distributions---into flow models.  During training, at each step, we construct causally-ordered sequences by sampling multiple images from the same semantic category and applying different levels of noise, where images with higher noise levels serve as causal predecessors to those with lower noise levels. This design enables the model to learn broader category-level variations while maintaining proper causal relationships in the flow process.  During generation, the model autoregressively conditions the previously generated images from earlier denoising steps, forming a contextual and coherent generation trajectory. 
Additionally, we design a customized hybrid linear attention mechanism tailored to our modeling approach to enhance computational efficiency.
Our approach, termed ARFlow, achieves 6.63 FID scores on ImageNet at 256 × 256 without classifier-free guidance, reaching 1.96 FID with classifier-free guidance 1.5, outperforming the previous flow-based model SiT's 2.06 FID. Extensive ablation studies demonstrate the effectiveness of our modeling strategy and chunk-wise attention design. 

\end{abstract}   
\section{Introduction}
\label{sec:intro}

Image generation has witnessed rapid advancements in recent years, with flow models~\cite{albergo2022building, lipman2022flow, liu2022flow, ma2024sit, esser2024scaling} emerging as a competitive solution. Compared to the curved trajectories observed in diffusion models~\cite{ho2020denoising, sohl2015deep, song2020score}, flow models connect data and noise through straight-line paths, resulting in higher training and inference efficiency. However, current flow-based approaches~\cite{gu2024dart} face a critical limitation: they struggle to capture long-range dependencies across the generation process. This limitation arises because each generation step has access only to the corrupted image from the immediately preceding step, forcing the model to compress all historical information into a single noisy intermediate state. Consequently, this constraint potentially degrades the model's ability to maintain semantic consistency and structural coherence throughout the generation process.

In contrast to flow models that rely on Markovian assumptions, autoregressive models naturally maintain historical information through their hidden states; that is, in autoregressive modeling, each token's prediction explicitly depends on all previous tokens. As demonstrated in modern large language models, such explicit modeling remains valid and powerful for building global relationships, even with sequences consisting of thousands or even millions of tokens~\cite{team2024gemini, lieber2024jamba}. This powerful capability suggests a promising solution for flow models—if integrated properly, it allows flow models to directly optimize for consistency across the generation trajectory while maintaining their computational benefits.

To this end, we develop \textit{ARFlow}, a novel framework that integrates autoregressive modeling within a flow-based architecture. Our key insight is to leverage the natural ordering inherent in the flow process to establish causal relationships during training. 
Empirically, by observing that directly concatenating sequences from flow trajectories leads to model collapse during training, we alternatively propose sampling multiple images from the same semantic category to form a sequence. This training strategy departs from the traditional single-image trajectory, enabling the model to learn broader semantic relationships while avoiding convergence to trivial solutions that merely capture differences between flow states.
Moreover, unlike the original flow models, we apply varying levels of independent noise to these sampled images, arranging them in a sequence---where more heavily corrupted images precede those with less noise---to help establish a clear causal relationship in the denoising process.
At the inference stage, ARFlow operates autoregressively: at each step, the model conditions on the complete sequence of previously generated images while following the flow dynamics to predict the next image. This autoregressive behavior allows the model to capture both local transitions through flow-based updates and long-range dependencies through the sequential memory of past states, resulting in more coherent and stable generation trajectories.

Furthermore, considering that ARFlow is expected to generate long sequences of image tokens---especially during autoregressive inference---we introduce a carefully designed linear attention mechanism as a replacement for self-attention, ensuring that the framework remains computationally feasible and scalable.  
Specifically, since the autoregressive relationships in ARFlow are modeled at the image level rather than the patch level, we propose a tailored hybrid linear attention mechanism that groups tokens into chunks corresponding to different images. 
By treating each image as a chunk, this mechanism implements full attention within chunks to model the generation of a single image, while applying causal masking only to inter-chunk attention to model image-wise relationships. This design preserves the necessary causality introduced by ARFlow and enables efficient parallel processing.
Extensive experiments on ImageNet are provided to demonstrate the effectiveness of ARFlow. For example, ARFlow achieves an FID of 4.34 with classifier-free guidance of 1.5, significantly outperforming the flow model SiT (FID 9.17) on 400k training steps on resolution of $128 \times 128$, and achieves FID of 1.96 outperforming the SiT(2.06) on 7M training steps on resolution of $256 \times 256$. 
Moreover, our ablation studies further demonstrate three key strengths of our framework: 1) longer sequence lengths consistently enhance generation quality, with FID improving from 29.12 to 25.01 as sequences increase from 2 to 10 steps; 2) the cached state mechanism proves essential, as its removal results in a FID degradation from 25.46 to 65.33; and 3) our hybrid attention design can efficiently process long autoregressive sequences, rendering ARFlow's generation speed comparable to SiT.
\section{Related Works}

\subsection{Generative Image Modeling}
 Diffusion models~\cite{sohl2015deep,ho2020denoising,song2020score,rombach2022high} have first garnered significant attention for their ability to generate high-quality, realistic images. Recent advancements in diffusion models have primarily been driven by innovations in sampling techniques~\cite{ho2020denoising,karras2022elucidating,song2020denoising}, model architectures such as DiT~\cite{peebles2023scalable} and U-ViT~\cite{bao2023all}, and the reformulation of the model output to represent velocity~\cite{salimans2022progressive} or noise~\cite{ho2020denoising} rather than pixels. 
More recently, flow models~\cite{lipman2022flow,liu2022flow,albergo2022building,ma2024sit} attracts increasing attention. Unlike diffusion models that follow curved trajectories, flow models connect data and noise distributions through straight-line paths, offering a more direct and efficient approach to generative modeling. 
However, these approaches are inherently limited by their Markovian nature: each generation step can only access information from the immediate previous step, constraining their ability to maintain long-range consistency.


 Autoregressive image generation models~\cite{chen2020generative,van2016conditional} begin by generating pixels sequentially. VQGAN~\cite{esser2021taming} enhances these models by employing autoregressive learning within the latent space of VQVAE~\cite{van2017neural}. Parti~\cite{yu2022scaling}, utilizing the ViT-VQGAN~\cite{yu2021vector} architecture, scales this approach further with a transformer consisting of 20 billion parameters. LlamaGen~\cite{sun2024autoregressive} use vanilla llama~\cite{touvron2023llama} for image generation.However, these models face a fundamental challenge: they must convert 2D images into 1D sequences through raster-scan ordering, limiting their ability to capture global structure. 
 VAR~\cite{tian2024visual} attempts to address this by introducing multi-scale token maps generated from coarse to fine scales, but struggles with error accumulation when predicting multiple tokens in parallel.

There are prior efforts in combining diffusion model and autoregressive model. MAR~\cite{li2024autoregressive} integrates the Mask generative model with a diffusion model, employing a transformer with bi-directional attention and using an MLP as the diffusion model to sample image latent conditioned on the transformer's output. DART~\cite{gu2024dart} organizes the denoising process trajectories of the diffusion model into a sequence,  reducing the number of diffusion steps from 1000 to 16 or 4. However, this reduction compromises the generative quality at each step and limits the autoregressive model's ability to fully leverage its long-distance modeling capabilities.

In contrast to these approaches, our \textit{ARFlow} maintains the advantages of standard flow steps while enhancing them with autoregressive conditioning.

\subsection{Linear Attention and Hybrid Model}
Transformers with self-attention mechanisms have been the backbone of foundation models  but struggle with quadratic training complexity and costly KV cache management during inference.
Linear attention~\cite{katharopoulos2020transformers}, which replaces the exponential similarity function (i.e., $\operatorname{softmax}$) with a dot product over (transformed) query and key vectors, has gained attention due to its hardware-efficient, sub-quadratic training~\cite{yang2023gated} and constant-memory recurrent inference capabilities (viewed as an RNN). However, linear attention suffers from notable performance degradation. To address this, researchers have revisited ideas from RNN literature, introducing data-independent static decay terms~\cite{sun2023retentive} or data-dependent decay terms~\cite{yang2023gated, qin2024hgrn2, choumetala, zhang2024gated, sun2024you}, which have improved linear attention’s performance. Notably, Mamba2~\cite{dao2024transformers} can be viewed as a special case of linear attention as denoted by \cite{yang2024parallelizing}, where data-dependent decay is crucial for both Mamba~\cite{gu2024mamba} and Mamba2~\cite{dao2024transformers}.

However, linear attention’s expressive power is still limited by the removal of the $\operatorname{softmax}$ operator. Researchers are exploring hybrid approaches, combining $\operatorname{softmax}$-based chunk or sliding window attention (we refer to them as local attention, which is sub-quadratic in total sequence length once window/chunk size is fixed) with linear attention. Models like Samba \cite{Ren2024-hz} interleave local attention and linear attention across different layers, while Gated Attention Unit (GAU)~\cite{hua22gau} and Infini-Attention~\cite{infiniattn} combine local attention  within a single layer. Here, the intra-chunk component uses  (local) $\operatorname{softmax}$ attention, while the inter-chunk component relies on (global) linear attention, allowing for more detailed local context modeling via $\operatorname{softmax}$ and access to global context through linear attention.

Our proposed hybrid linear attention method aligns more closely with GAU~\cite{hua22gau} and Infini-Attention~\cite{infiniattn}. 
We treat a chunk as a patchified image sequence at a given time step in the flow process, applying $\operatorname{softmax}$ attention to model interactions between patches within each image, while using linear attention to compress intermediate images from all preceding flow steps, facilitating global information exchange. 
To suit flow models' properties, we modify intra-chunk $\operatorname{softmax}$ causal attention to bidirectional attention.
  
\section{Preliminary}
\begin{figure*}[htbp]
\centering
\includegraphics[width=0.83\linewidth]{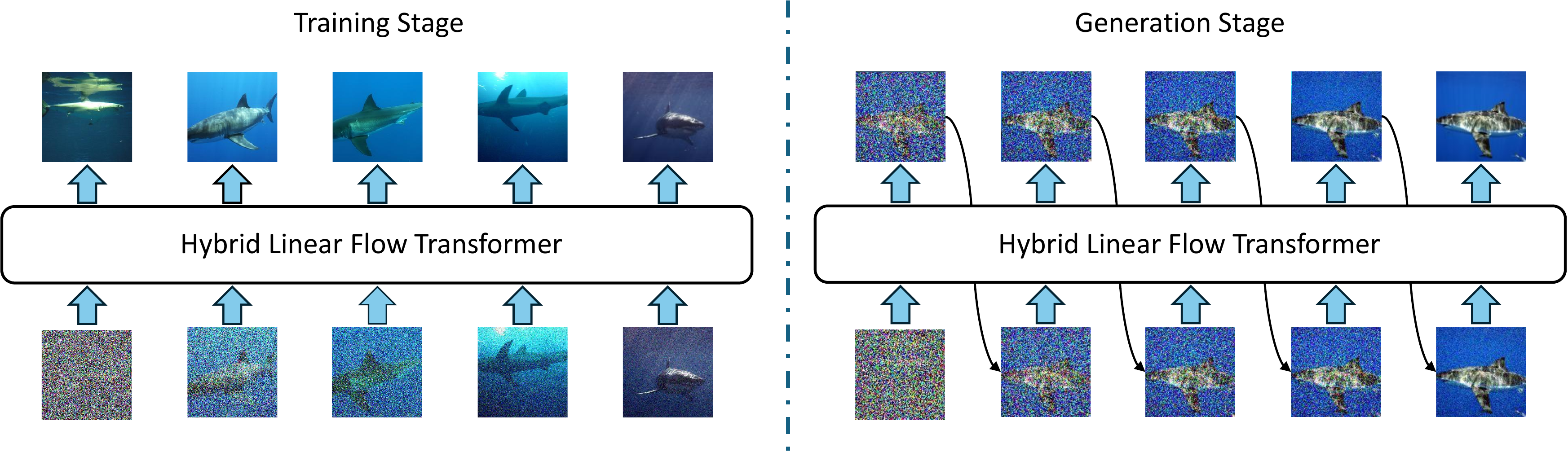}

\caption{Overview of ARFlow's training and generation stages. Left: During training, the model processes multiple images from the same semantic category with different noise levels through a Hybrid Linear Flow Transformer, enabling learning of broader category-level variations. The samples are arranged in order of increasing noise levels to establish causal relationships. Right: During generation, the model autoregressively conditions on all previously generated images (indicated by curved arrows), creating a coherent generation trajectory from noise to the final image.}
\label{fig:model}
\vspace{-3mm}
\end{figure*}


\subsection{Latent generative models}
Training generative models directly in high-resolution pixel space presents significant computational challenges, particularly when dealing with modern image resolutions. Recent works~\cite{esser2021taming,rombach2022high,li2024autoregressive} have adopted a two-stage approach that has proven highly effective. This approach introduces an intermediate latent space that dramatically reduces the computational burden while preserving image quality.

The first stage involves training an autoencoder (VQ or KL-based), where the encoder $E$ maps a high-dimensional raw image $\mathbf{X^*} \in \mathbb{R}^{3 \times H \times W}$ into a significantly more compact latent representation $\mathbf{Z^*} = E(\mathbf{X^*}) \in \mathbb{R}^{d \times h \times w}$:
where the dimensionality reduction is substantial: $3\times H\times W \gg d\times h\times w$. This compression typically reduces the spatial dimensions by a factor of 8 or more while also reducing the channel dimension.

In the second stage, the generative model is trained exclusively in this lower-dimensional latent space, with the autoencoder's weights remaining frozen. This separation of concerns allows the autoencoder to focus on learning an efficient compression scheme, while the generative model can concentrate on modeling the underlying data distribution in a more tractable space. During generation, the model produces samples in the latent space, which are then transformed back into pixel space using the decoder $D$. Unless explicitly stated otherwise, our subsequent discussion focuses on operations within this latent space.

\subsection{Flow models}
Flow models represent a powerful approach to generative modeling by leveraging continuous-time stochastic processes. Unlike diffusion models that follow curved paths between data and noise distributions, flow models create straight-line trajectories, leading to more efficient training and sampling. These models operate by gradually transforming standard Gaussian noise $\epsilon \sim \mathcal{N}(0, \mathbf{I})
$ into image latents $\mathbf{Z^*} \sim p(\mathbf{Z})$ through a carefully designed process:
\begin{align} 
\mathbf{Z}_t = \alpha_t \mathbf{Z^*} + \sigma_t \eps, 
\label{eq2} 
\end{align}
where $t \in [0,1]$ represents the continuous time parameter. The scheduling functions $\alpha_t$ and $\sigma_t$ play crucial roles:
$\alpha_t$ decreases monotonically from 1 to 0 as $t$ increases, controlling the contribution of the target image, while $\sigma_t$ increases from 0 to 1, gradually introducing noise.  common choice is linear scheduling where:
\begin{equation}
\alpha_t = 1-t, \quad \sigma_t = t
\end{equation}
This creates a smooth interpolation between the clean image latent and pure noise, allowing for stable training and generation.
Flow models with parameters $\theta$ in the velocity field can be optimized by minimizing:
\begin{align}
    \label{eq:velocity-eq-obj}
    \LL_{\mathrm{v}}(\theta) 
    &= \int_0^T \E[\Vert \bv_\theta(\mathbf{Z}_t, t) - \dot\alpha_t \mathbf{Z}^* - \dot\sigma_t \eps\Vert^2] \rd t .
\end{align}
Here, $v_\theta$ predicts the instantaneous change in $Z_t$, and the objective ensures this prediction matches the ground truth velocity given by $\dot{\alpha}_t Z^* + \dot{\sigma}_t \varepsilon$.
This objective ensures accurate prediction of instantaneous changes in the latent representation across the entire trajectory, enabling precise control over the generation process.

\subsection{Chunkwise linear attention} Denote query/key /value matrix as $\rmQ, \rmK, \rmV \in \mathbb{R}^{T\times d}$ where $T$ is sequence length and $d$ is head dimension,  for each head,
  the (simplified) linear attention has the following two equivalent recurrent and parallel form \cite{yang2023gated}:
\begin{align}
\hspace{-3mm}\mathbf{S}_{t} &= \mathbf{S}_{t-1} + \mathbf{k}_{t}\mathbf{v}_{t}^{\top} \in \mathbb{R}^{d\times d}, \mathbf{o}_{t} = \mathbf{S}_{t}^\top\mathbf{q}_{t} \in \mathbb{R}^{d} \tag{recurrent} \\
\hspace{-3mm}
\mathbf{O} &= \left(\mathbf{Q}\mathbf{K}^\top \odot \mathbf{M}\right) \mathbf{V} \quad \in \mathbb{R}^{T \times d}, \tag{parallel}
\end{align}
where $\mathbf{M}\in \{0,1\}^{T\times T}$ is the causal mask.
Another equivalent chunk-wise parallel form balances between parallel and recurrent form~\cite{GAU,sun2023retentive,yang2023gated}, enabling subquadratic hardware-efficient training~\cite{yang2023gated}. Consider an input sequence $\mathbf{X} \in \mathbb{R}^{T \times d}$ divided into non-overlapping chunks of length $C$ each, where $T = N C$ and $N$ represents the number of chunks. Let $\mathbf{S}_{[i]} := \mathbf{S}_{iC} \in \mathbb{R}^{d \times d}$ denote the hidden state after processing $i$ chunks (i.e., after time step $t = iC$). Define $\mathbf{Q}_{[i+1]} = \mathbf{Q}_{iC+1:iC+C} \in \mathbb{R}^{C \times d}$ as the query vectors for the $(i+1)$-th chunk, with analogous definitions for $\mathbf{K}_{[i+1]}, \mathbf{V}_{[i+1]}, \mathbf{O}_{[i+1]}$.
Then the \emph{chunk-level} recurrence could be written as:
\begin{align}
\mathbf{S}_{[i+1]} &= \mathbf{S}_{[i]} + \sum_{t = iC+1}^{iC+C} \mathbf{k}_t \mathbf{v}_t^\top 
= \mathbf{S}_{[i]} + \mathbf{K}_{[i+1]}^\top \mathbf{V}_{[i+1]} \in \mathbb{R}^{d \times d}.\label{eq:recurrence}
\end{align}
and the intra-chunk (parallel) output computation is :
\begin{align}
\mathbf{O}_{[i+1]} &= \underbrace{\mathbf{Q}_{[i+1]} \mathbf{S}_{[i]}}_{\text{inter-chunk}} +\underbrace{\left( (\mathbf{Q}_{[i+1]} \mathbf{K}_{[i+1]}^\top) \odot \mathbf{M} \right) \mathbf{V}_{[i+1]}}_{\text{intra-chunk}} \in \mathbb{R}^{C\times d}
\end{align}
The intra-chunk component $\mathbf{O}^{\text{intra}}_{[i+1]}$ requires $O(C^2 d + C d^2)$ computational time, and the inter-chunk component $\mathbf{O}^{\text{inter}}_{[i+1]}$ processes contributions from previous chunks' hidden states in $O(C d^2)$ time. Once $C$ is fixed to a small constant (e.g., 64 or 128 in practice), the overall training time is linear in sequence length $T$. Notably, the output computation is highly parallelizable and the chunk-level recurrence significantly reduce the recurrent step compared to the recurrent form, enabling hardware-efficient training \cite{yang2023gated,yang2024fla}.


\section{ARFlow with Hybrid Linear Attention}
\subsection{ARFlow}
The straightforward approach of flow model, however, suffers from two fundamental limitations. 
First, the entire sequence shares the same source image, providing insufficient variation for learning meaningful representations. 
Second, while preserving Markov chain properties, this configuration often leads to degenerate solutions that merely interpolate between noise levels, failing to capture broader semantic relationships.

To address these limitations, we propose a novel non-Markovian sequence construction method that introduces strategic variation in both content ($\mathbf{Z}^*$) and noise level ($\sigma$). 
Our approach consists of three key steps:   

First, we sample $N$ images $\mathbf{X}_n^*$ from the same semantic category with condition $c$, alongside $N$ independently sampled time values $t_n \in [0, 1]$. 
This sampling strategy ensures diverse content while maintaining semantic consistency within each sequence.

Second, we transform these images into latent representations $\mathbf{Z}_n^*$ through our encoder (Equation 1), then perturb them with independently sampled noise $\boldsymbol{\varepsilon}_n$ according to the flow process (Equation 2). 
The independence of noise samples further enhances the diversity of the learning signal.
Finally, we arrange the sequence based on the temporal ordering of $t_n$:
\begin{equation}
    \text{Seq}_N = [\mathbf{Z}_{t_1}^1, \mathbf{Z}_{t_2}^2, ..., \mathbf{Z}_{t_N}^N]
    \label{eq:img_seq}
\end{equation}
This enriched sequence forms the basis for our AR flow model optimization:
\begin{equation}
    \mathcal{L}_\theta = \int_0^T \sum_{n=1}^N \mathbb{E}[\|v_\theta(\mathbf{Z}_{t_n}^n, ..., \mathbf{Z}_{t_0}^0, t) - \dot{\alpha}_t \mathbf{Z}_n^* - \dot{\sigma}_t \boldsymbol{\varepsilon}_n\|^2]dt
\end{equation}
During the generation phase, we implement a next-flow prediction mechanism, where each autoregressive step corresponds to a complete flow sampling step for an entire latent image $\mathbf{Z}$, rather than individual tokens $z$. 
This design choice maintains efficiency while preserving the benefits of autoregressive modeling.

Our approach offers several compelling advantages: By incorporating multiple images from the same category, the model learns to capture rich semantic relationships and category-level variations.The independent sampling of both images and flow times creates diverse patterns that prevent convergence to trivial solutions. While breaking the restrictive Markovian assumption, our approach maintains the progressive nature of flow models, ensuring temporal coherence. The model effectively leverages information from multiple previous steps while maintaining computational tractability.

\subsection{Hybrid linear attention}

In practice, to obtain a sequence, for each latent image $\mathbf{Z}_{t_i}^i$ (Eq.~\ref{eq:img_seq}) we patchify it into $M$ patches in raster-scan order, i.e., sweeping them from left to right and top to bottom, resulting in $\mathbf{Z}^{i}_{t_i,m} \in \mathbb{R}^{d}$ where $m \in \{1, \cdots M\}$ (typically $M=h\times w=64$ when image size is 128 $\times$ 128). 
Then we concatenate them to a sequence \(
\mathbf{X} = \{ \mathbf{Z}^{1}_{t_1, 1}, \cdots \mathbf{Z}^{1}_{t_1, M}, \mathbf{Z}^{2}_{t_2, 1}, \cdots \mathbf{Z}^{N}_{t_N, M} \}
\). 
Similar to standard language modeling, we apply query/key/value linear projection to obtain $\rmQ, \rmK, \rmV \in \mathbb{R}^{T \times d}$ where $T=N\times M$.

In practice, however, $T$ can be very large, making Transformer-based autoregressive modeling expensive, particularly during inference due to the large KV cache size. To address this, we adopt linear attention to reduce complexity and make necessary adaptations to suit flow models.

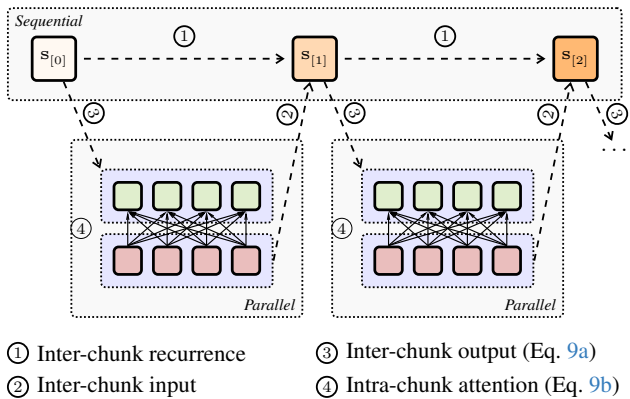
\begin{wrapfigure}{r}{0.45\textwidth}
\centering
\vspace{-3mm}
\resizebox{\linewidth}{!}{
  \centering
  \begin{tikzpicture}[
    qnode/.style={
        draw,
        minimum width=12pt,
        minimum height=12pt,
        very thick,
        rounded corners=2pt,
        fill=blue!15,
        font=\tiny,
      },
    qkvnode/.style={
        draw,
        densely dotted,
        thick,
        inner sep=2pt,
        minimum width=19pt,
        minimum height=19pt,
        rounded corners=2pt,
        font=\tiny,
      },
    attnnode/.style={
        draw,
        densely dotted,
        thick,
        inner sep=2pt,
        minimum width=19pt,
        minimum height=19pt,
        rounded corners=2pt,
        font=\tiny,
      },
    hnode/.style={
        draw,
        densely dotted,
        thick,
        rounded corners=2pt,
        font=\tiny,
      },
    hidden/.style={
        draw,
        minimum width=19pt,
        minimum height=19pt,
        very thick,
        rounded corners=2pt,
        font=\tiny,
      },
    hiddenlink/.style={
        dashed,
        ->,>={Straight Barb[width=4pt]},
        thick,
        shorten >=3pt,
        shorten <=3pt,
    },
    olink/.style={
        ->,>={Straight Barb[width=4pt]},
        dashed,
        thick,
    },
    alink/.style={
        ->,>={Straight Barb[width=1pt]},
        line width=0.5pt,
    },
    ]

    \foreach \i in {1,...,2}{
        \node[qnode,anchor=west,fill=limegreen!30] (i\i) at (\i*4+0.9,0) {};
        \node[qnode,anchor=west,fill=limegreen!30] (q1\i) at ([xshift=4pt]i\i.east) {};
        \node[qnode,anchor=west,fill=limegreen!30] (q2\i) at ([xshift=4pt]q1\i.east) {};
        \node[qnode,anchor=west,fill=limegreen!30] (q3\i) at ([xshift=4pt]q2\i.east) {};
        
        \node[qnode,anchor=north,fill=brickred!30] (q4\i) at ([yshift=-15pt]i\i.south) {};
        \node[qnode,anchor=north,fill=brickred!30] (q5\i) at ([yshift=-15pt]q1\i.south) {};
        \node[qnode,anchor=north,fill=brickred!30] (q6\i) at ([yshift=-15pt]q2\i.south) {};
        \node[qnode,anchor=north,fill=brickred!30] (q7\i) at ([yshift=-15pt]q3\i.south) {};
        
        \begin{scope}[on background layer]
        \node[attnnode, fit=(i\i)(q7\i), inner sep=18pt, fill=gray!5] (attn\i) {};
        \node[anchor=south east] at (attn\i.south east) {\scriptsize\emph{Parallel}};
        \end{scope}
        \begin{scope}[on background layer]
        \node[qkvnode, fit=(i\i)(q3\i), inner sep=5pt, fill=blue!10] (qkv0\i) {};
        \node[qkvnode, fit=(q4\i)(q7\i), inner sep=5pt, fill=blue!10] (qkv1\i) {};
        \end{scope}

        \foreach \j in {4,...,7}{
            \draw[alink] (q\j\i.north) -- (i\i.south);
            \foreach \k in {1,...,3}{
                \draw[alink] (q\j\i.north) -- (q\k\i.south);
            }
        }
        \node[circle,draw=black, anchor=west,inner sep=1pt] at (attn\i.west) {\scriptsize$4$};
    }

    \node[] at ($(attn2.east)+(0.8,1.2)$) (input1) {$\dots$};
    
    \foreach \i in {1,...,3}{
    \tikzmath{\j=int(\i-1); }
    \node[hidden, fill=orange!\the\numexpr\i*25-20] (h0\i) at (\i*4,60pt) {$\mathbf{S}_{[\j]}$};
    \ifnum\i>1
      \draw[hiddenlink] (h0\the\numexpr\i-1) -- (h0\i) node[circle,midway,above=5pt,draw, solid, inner sep=1pt] {\scriptsize$1$};
      \draw[olink] (qkv1\the\numexpr\i-1.east) -- (h0\i) node[circle,midway,above=2pt,pos=0.8,sloped,draw, solid, inner sep=1pt] {\scriptsize$2$};
      \draw[olink] (h0\the\numexpr\i-1) -- (qkv0\the\numexpr\i-1.north west) node[circle,midway,above=2pt,pos=0.43,sloped,draw, solid, inner sep=1pt] {\scriptsize$3$};
    \fi
    }
    \draw[olink] (h03) -- (input1) node[circle,midway,above=2pt,pos=0.65,sloped,draw, solid, inner sep=1pt] {\scriptsize$3$};

    \begin{scope}[on background layer]
    \node[hnode, fit=(h01)(h03), inner sep=10pt, yshift=2pt, fill=gray!5] (h) {};
    \node[anchor=north west] at (h.north west) {\scriptsize\emph{Sequential}};
    \end{scope}
    
    \node[circle,draw=black, thick, anchor=west,inner sep=1pt] (c1) at ($(h.south west)+(0,-110pt)$) {\scriptsize$1$};
    \node[circle,draw=black, thick, inner sep=1pt] (c2) at ($(c1.south)+(0,-10pt)$) {\scriptsize$2$};
    \node[circle,draw=black, thick, inner sep=1pt] (c3) at ($(c1.east)+(130pt,0pt)$) {\scriptsize$3$};
    \node[circle,draw=black, thick, inner sep=1pt] (c4) at ($(c3.south)+(0,-10pt)$) {\scriptsize$4$};
    
    \node[anchor=west] at ($(c1.east)+(0pt,0pt)$) {Inter-chunk recurrence};
    \node[anchor=west] at ($(c2.east)+(0pt,0pt)$) {Inter-chunk input};
    \node[anchor=west] at ($(c3.east)+(0pt,0pt)$) {Inter-chunk output (Eq.~\ref{eq:inter-attn})};
    \node[anchor=west] at ($(c4.east)+(0pt,0pt)$) {Intra-chunk attention (Eq.~\ref{eq:intra-attn})};
  \end{tikzpicture}

}
\caption{Architecture of our hybrid chunk-wise attention mechanism. The computation combines sequential processing between chunks (shown in the upper ``Sequential'' section) with parallel processing within chunks (shown in the lower ``Parallel'' section). Numbers indicate key operations: (1) chunk-level state updates with decay, (2) information aggregation from current chunk, (3) inter-chunk attention through hidden states, and (4) full attention within each chunk. This design enables efficient processing of long sequences while maintaining appropriate causal structure for flow modeling.}
\label{fig:chunk-attn}
\vspace{-16mm}
\end{wrapfigure}
Unlike in language modeling, flow models do not require enforcing causality among tokens (or patches) within each image. Instead, modeling bidirectional interactions enhances the coherence and quality of generated images. To leverage this, we propose a hybrid linear attention mechanism for ARFlow, partitioning the sequence into chunks, each representing a separate image, i.e., $C = M$. As shown in Figure~\ref{fig:chunk-attn}, our hybrid linear attention retains sequential dependencies via hidden states while allowing parallel processing across 
 different chunks, like chunkwise linear attention introduced in. 
 The autoregressive flow structure aligns naturally with this chunk-based linear attention formulation. Unlike next-token prediction, flow models focus on next de-noising image prediction, supporting full attention within chunks while preserving causal dependencies between them. For the $(i+1)$-th chunk, the computation is formulated as:
\begin{subequations}
\begin{align}
\mathbf{O}_{[i+1]} &= \underbrace{\mathbf{Q}_{[i+1]} \mathbf{S}_{[i]}}_{\text{inter-chunk (causal)}}  \label{eq:inter-attn}\\
&+ \underbrace{\operatorname{softmax}\left(\mathbf{Q}_{[i+1]} \mathbf{K}_{[i+1]}^\top\right) \mathbf{V}_{[i+1]}}_{\text{intra-chunk (local noncausal attention)}}, \label{eq:intra-attn}
\end{align}
\end{subequations}
where the inter-chunk component maintains causality through the sequential hidden state $\mathbf{S}_{iC}$, while the intra-chunk component implements unrestricted attention within the current chunk. The hidden state update remains intact. 
\paragraph{Incorporating gating mechanism.}
Gating mechanism or selective mechanism has been commonly used to enhance linear attention \cite{yang2023gated,zhang2024gated, sun2024you} or state space models \cite{gu2024mamba,dao2024transformers}. Empirically, we found that scalar-valued decay term \cite{gu2024mamba, sun2024you} suffices to produce good performance while enjoying faster training speed, and we make the decay term data-dependent:
\begin{equation}
g_t = \operatorname{sigmoid}\left( \mathbf{W}_\gamma   \boldsymbol{x}_t \right)^{1/\tau} \in (0, 1)
\end{equation}
where $\boldsymbol{x}_t\in \mathbf{X}$ is the input for each step, and $\tau=16$ is the temperature used for regulating the decaying factor to $1$ \cite{yang2023gated, sun2024you} for retaining more historical information.
Within each chunk $(i+1)$, we compute the geometric mean of the gating values to obtain the \emph{chunk-level} decay:
\begin{align}
\gamma_{i+1} &= \exp\left( \frac{1}{C} \sum_{t=iC+1}^{iC+C} \log g_t \right) 
\end{align}

\begin{equation}
\rmS_{[i+1]} =\gamma_{i+1} \rmS_{[i]} + \left( \\{\rmK}_{[i+1]}^{\top} \rmV_{[i+1]} \right)
\end{equation}
and output computation remains intact.


\section{Experiments}
\subsection{Experiment Setup}

 \paragraph{Training.} We train a class-conditional latent model on the ImageNet dataset~\cite{deng2009imagenet} $128 \times 128$ and $256 \times 256$,  a highly-competitive generative modeling benchmark. Following the original DiT~\cite{peebles2023scalable}, we initialize the final linear layer with zeros and use standard weight initialization techniques from ViT. All models are trained using the AdamW optimizer with a constant learning rate of $1 \times 10^{-4}$, no weight decay, and a batch size of 256. Unless otherwise noted, the sequence length is set to 5. We maintain an exponential moving average (EMA) of model weights with a decay rate of 0.9999, and all reported results use the EMA model. The same training hyperparameters are used across all models, including the SiT~\cite{ma2024sit} baselines.
\paragraph{Flow Setting}
We use the same pre-trained variational autoencoder (VAE) model from Stable Diffusion as employed in SiT~\cite{ma2024sit} to encode images, producing a representation of dimensions $32 \times 32 \times 4$ for input images of size $128 \times 128 \times 3$. A linear interpolation schedule is used for $t$, where $\alpha_t = 1 - t$ and $\sigma_t = t$ with $t \in[1, 0]$. During generation, we apply the SDE solver using the first-order Euler-Maruyama integrator~\cite{ma2024sit}, and limit the number of generation steps to 250, consistent with the sampling steps used in SiT. 
Unless otherwise specified, all metrics presented are FID-50K scores evaluated on the ImageNet dataset.
\subsection{Comparison with DiT Backbone Models}


Table~\ref{tab:backbone} presents a performance comparison of ARFlow, SiT, and DiT models on the ImageNet $256 \times 256$ dataset under varying classifier-free guidance (Cfg) conditions.
Without classifier-free guidance, ARFlow consistently outperforms DiT and SiT, achieving notably superior results in IS (133.5), FID (6.63), and sFID (5.17). Under classifier-free guidance, ARFlow further extends its advantage, demonstrating the highest IS (288.4) alongside the lowest FID (1.96) and sFID (4.17). These results clearly indicate ARFlow's superior capability in generating high-quality images compared to the baseline models, irrespective of guidance settings.
\begin{table}
\vspace{-6mm}
\caption{Performance comparison between ARFlow, SiT and DiT on ImageNet 256×256 across model scales and classifier-free guidance values. \textbf{Bold: The best.}}
\label{tab:backbone}
\resizebox{\linewidth}{!}{%
\begin{tabular}{ccccccccc}
    \toprule
    Model & Model Size & Steps & Cfg  & IS$\uparrow$ & FID$\downarrow$ & sFID$\downarrow$ & Precision$\uparrow$ & Recall$\uparrow$ \\
    \midrule
    DiT & XL/2  &7M& w/o & 121.5  & 9.62 & 6.85 & 0.67  & 0.67  \\
    SiT & XL/2  &7M& w/o & 131.7  & 8.61 & 6.32 & 0.68  & 0.67  \\
    ARFlow & XL/2 &7M  & w/o & \textbf{133.5}  & \textbf{6.63} & \textbf{5.17} & \textbf{0.68}  & \textbf{0.65}  \\
    \midrule
    DiT & XL/2  &7M& w & 278.2  & 2.27 & 4.60 & 0.83  & 0.57  \\
    SiT & XL/2  &7M& w & 270.3  & 2.06 & 4.49 & 0.83  & 0.59  \\
    ARFlow & XL/2 & 7M  & w & \textbf{288.4}  & \textbf{1.96} & \textbf{4.17} & \textbf{0.81}  & \textbf{0.61}  \\
    \bottomrule
\end{tabular}%
}
\end{table}
\subsection{Comparison with SiT}
\vspace{-3mm}
\begin{table}[htbp]
\centering
\caption{Performance comparison between ARFlow and SiT on ImageNet 128×128 across model scales and classifier-free guidance values. \textbf{Bold: The best.}}

\begin{subtable}[t]{0.48\textwidth}
\centering
\caption{CFG = 1.0}
\label{tab:sit_cfg1.0}
\resizebox{\linewidth}{!}{%
\begin{tabular}{lccccc}
\toprule
Model & Size & IS & FID & Prec & Rec \\
\midrule
SiT     & S/2  & 18.45 & 49.21 & 0.3764 & 0.5847 \\
ARFlow  & S/2  & 22.15 & 41.56 & 0.4184 & 0.6060 \\
\midrule
SiT     & B/2  & 25.69 & 35.64 & 0.4539 & 0.6231 \\
ARFlow  & B/2  & 36.73 & 25.46 & 0.5185 & 0.6489 \\
\midrule
SiT     & L/2  & 35.68 & 25.39 & 0.5211 & 0.6488 \\
ARFlow  & L/2  & 55.03 & 15.91 & 0.5866 & 0.6651 \\
\midrule
SiT     & XL/2 & 37.59 & 24.02 & 0.5277 & 0.6481 \\
ARFlow  & XL/2 & \bf{60.78} & \bf{14.08} & \bf{0.5989} & \bf{0.6705} \\
\bottomrule
\end{tabular}%
}
\end{subtable}
\hfill
\begin{subtable}[t]{0.48\textwidth}
\centering
\caption{CFG = 1.5}
\label{tab:sit_cfg1.5}
\resizebox{\linewidth}{!}{%
\begin{tabular}{lccccc}
\toprule
Model & Size & IS & FID & Prec & Rec \\
\midrule
SiT     & S/2  & 31.84 & 31.13 & 0.4902 & 0.5530 \\
ARFlow  & S/2  & 41.48 & 23.95 & 0.5469 & 0.5779 \\
\midrule
SiT     & B/2  & 53.32 & 17.89 & 0.5984 & 0.5797 \\
ARFlow  & B/2  & 84.18 & 10.16 & 0.6607 & 0.5908 \\
\midrule
SiT     & L/2  & 79.90 & 10.05 & 0.6780 & 0.5759 \\
ARFlow  & L/2  & 130.12 & 4.92 & 0.7239 & 0.6020 \\
\midrule
SiT     & XL/2 & 85.10 & 9.17 & 0.6876 & 0.5755 \\
ARFlow  & XL/2 & \bf{140.07} & \bf{4.34} & \bf{0.7303} & \bf{0.6106} \\
\bottomrule
\end{tabular}%
}
\end{subtable}
\label{tab:SiT}
\end{table}

\paragraph{Metric Comparison} We compare ARFlow with SiT on ImageNet $128 \times 128$ across different model scales and classifier-free guidance (CFG) settings in~\ref{tab:SiT}. ARFlow consistently outperforms SiT across all model scales and metrics. Without classifier guidance (CFG=1.0), ARFlow-XL/2 achieves FID 14.08 and IS 60.78, substantially improving upon SiT-XL/2's 24.02 FID and 37.59 IS. With CFG=1.5(the improvements are more pronounced - ARFlow-XL/2 reaches 4.34  FID and 140.07 IS, marking a 52.67\% relative FID improvement over SiT-XL/2. Enhanced Precision/Recall metrics demonstrate improvements in both sample quality and diversity.

\begin{figure}[htbp]
\centering
\begin{subfigure}[b]{0.49\linewidth}
  \centering
  \includegraphics[width=\linewidth]{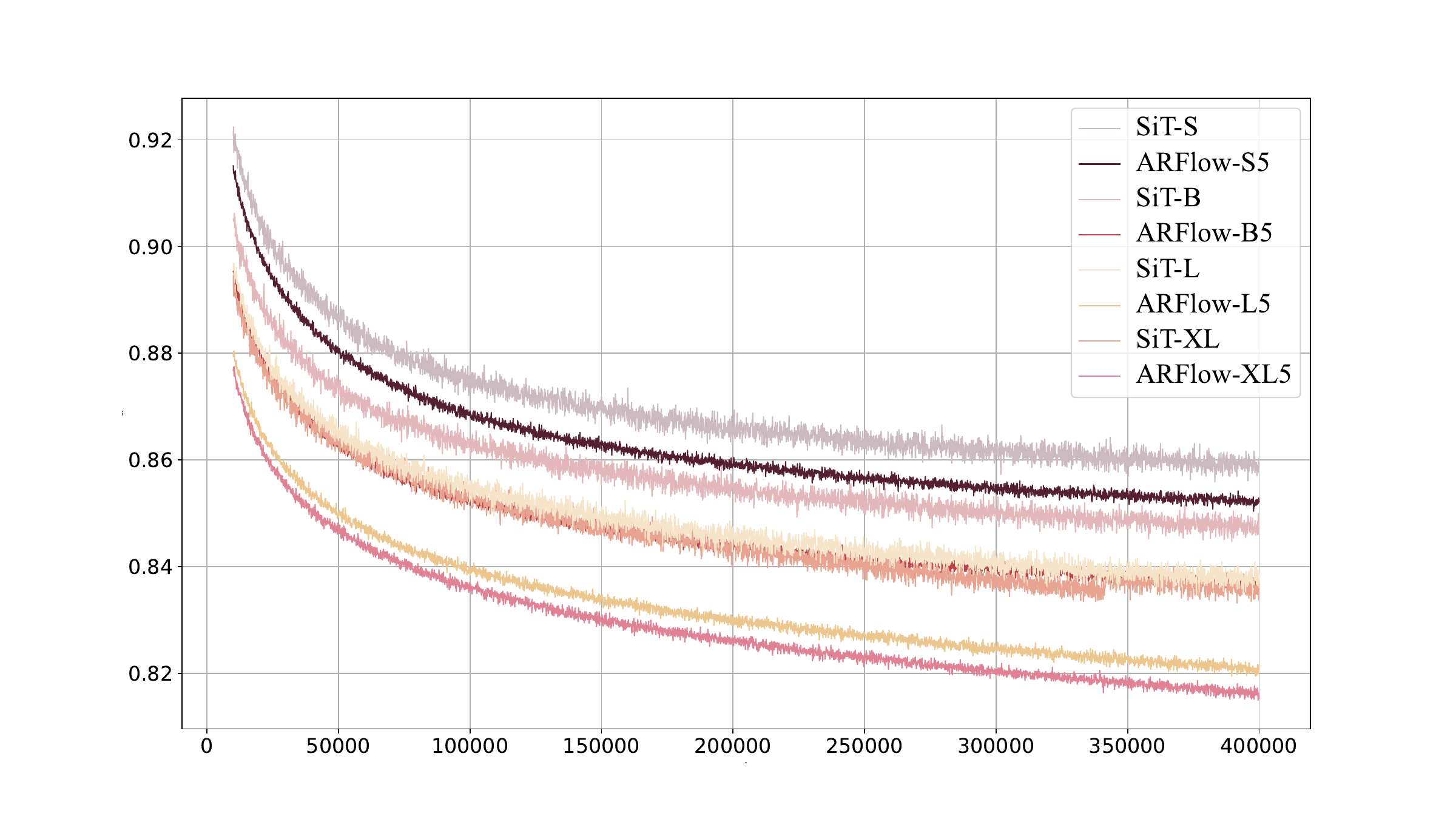}
  \caption{Training loss across different-scale models.}
  \label{fig:model_seq_lenth_loss}
\end{subfigure}
\hfill
\begin{subfigure}[b]{0.49\linewidth}
  \centering
  \includegraphics[width=\linewidth]{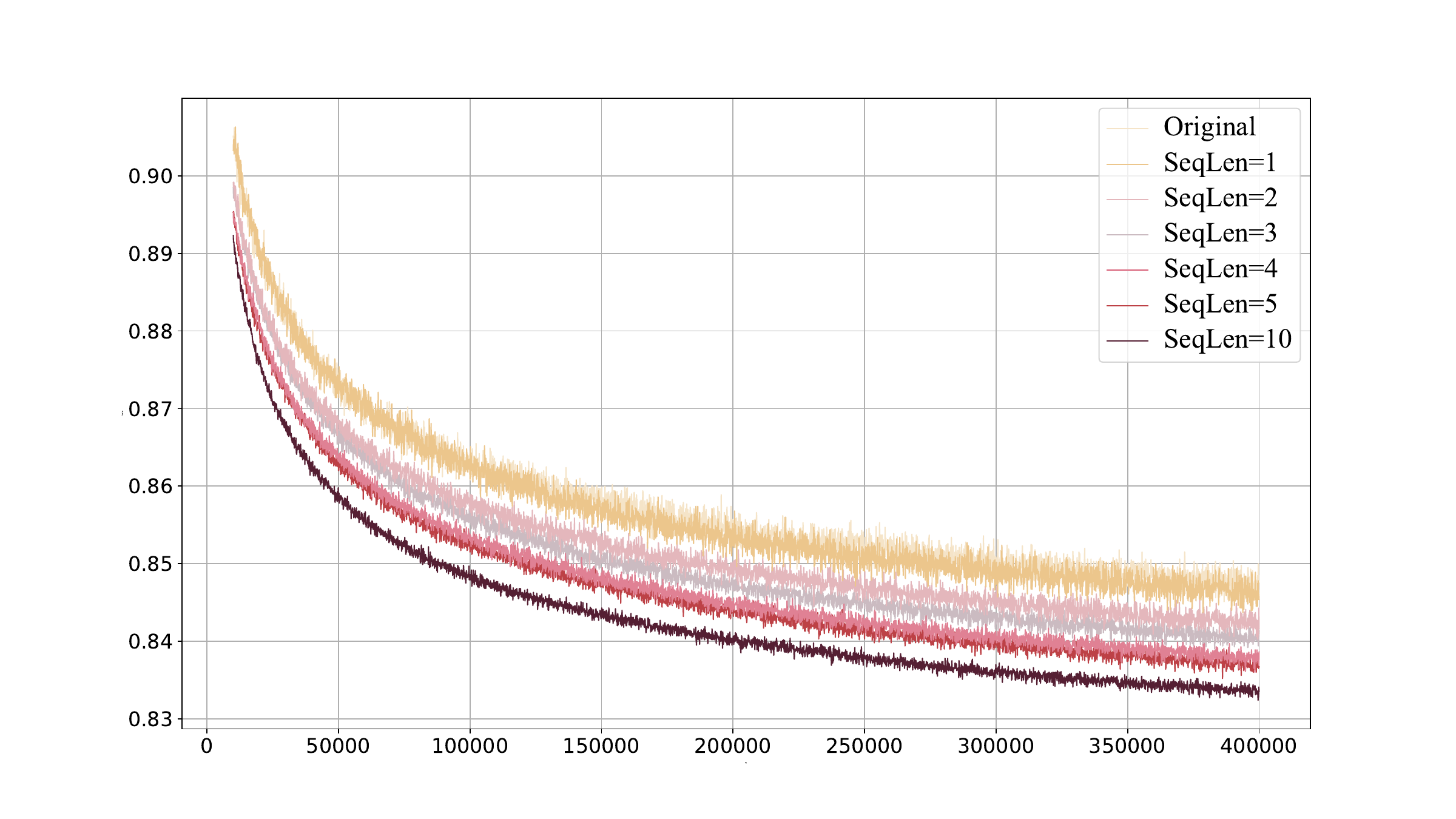}
  \caption{Training loss under varying sequence lengths.}
  \label{fig:model_seq_lenth}
\end{subfigure}
\vspace{-2mm}
\caption{Comparison of training loss between SiT and ARFlow across scale and sequence length.}
\label{fig:training_loss_combined}
\vspace{-4mm}
\end{figure}

\paragraph{Visualizing ARFlow Scale} As shows in Figure~\ref{fig:model_seq_lenth_loss},  we visualize the training loss of SIT and ARFlow for different scales, and we can find that for the same model size, ARFlow's loss is lower than that of the original SiT, and in particular, ARFlow-B5's loss is close to SiT-XL's loss.

\subsection{Ablation Study}
 We conduct extensive ablation studies using the ARFlow-B/2 as our baseline.
 \begin{figure}[htbp]
\centering
\includegraphics[width=\linewidth]{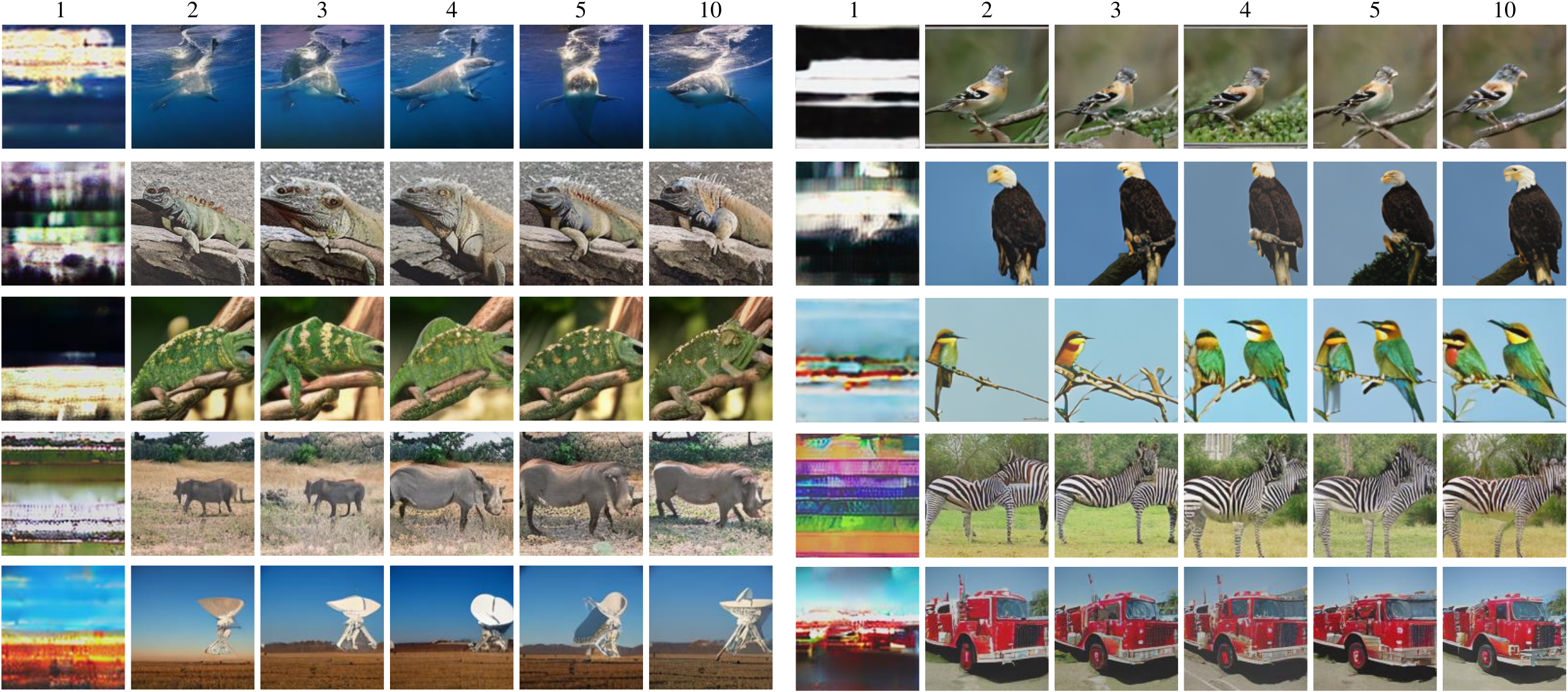}
\caption{Visual comparison of image generation quality across different training sequence lengths at ARFlow-B with CFG=4.0, demonstrating progressive improvement in sample quality as sequence length increases.}
\label{fig:smaples_seq_lenth}
\vspace{-1mm}
\end{figure}
\paragraph{Sequence length}

We first investigate the impact of sequence length on model performance in Table~\ref{table:seqlen}. With CFG=1.0, increasing the sequence length from 2 to 5 steps steadily improves the FID from 29.12 to 25.46. This trend continues up to 10 steps, reaching an FID of 25.01. Similar improvements are observed with CFG=1.5, where the FID decreases from 13.00 (2 steps) to 10.16 (5 steps) and further to 9.97 (10 steps). It is worth emphasizing that ARFlow training with sequence length of 1 cannot perform autoregressive flow generation, as it inherently reduces to the conventional flow model due to the absence of historical context required for autoregressive conditioning.

\vspace{-5mm}





\begin{table}[htbp]
\centering
\caption{Impact of sequence length on ARFlow-B/2 performance under different classifier-free guidance (CFG) strengths.}

\begin{adjustbox}{width=0.95\textwidth} 
\begin{tabular}{cc}
\begin{subtable}[t]{0.48\textwidth}
\centering
\footnotesize
\begin{tabular}{cccccc}
\toprule
Len & IS$\uparrow$ & FID$\downarrow$ & sFID & Prec & Rec \\
\midrule
1  & 5.23  & 141.14 & 50.42 & 0.1021 & 0.0730 \\
2  & 31.62 & 29.12  & 7.16  & 0.4959 & 0.6397 \\
3  & 33.22 & 27.66  & 7.15  & 0.5077 & 0.6469 \\
4  & 35.92 & 25.76  & 7.08  & 0.5173 & 0.6550 \\
5  & 36.73 & 25.46  & 7.13  & 0.5185 & 0.6489 \\
10 & 37.96 & 25.01  & 7.16  & 0.5189 & 0.6543 \\
\bottomrule
\end{tabular}
\caption{CFG = 1.0}
\end{subtable}
&
\begin{subtable}[t]{0.48\textwidth}
\centering
\footnotesize
\begin{tabular}{cccccc}
\toprule
Len & IS$\uparrow$ & FID$\downarrow$ & sFID & Prec & Rec \\
\midrule
1  & 5.50  & 135.71 & 47.13 & 0.1067 & 0.0251 \\
2  & 68.37 & 13.00  & 6.86  & 0.6429 & 0.5734 \\
3  & 73.14 & 12.02  & 6.79  & 0.6472 & 0.5796 \\
4  & 81.01 & 10.57  & 6.65  & 0.6581 & 0.5903 \\
5  & 84.18 & 10.16  & 6.77  & 0.6607 & 0.5908 \\
10 & 86.43 & 9.97   & 6.51  & 0.6629 & 0.5966 \\
\bottomrule
\end{tabular}
\caption{CFG = 1.5}
\end{subtable}
\end{tabular}
\end{adjustbox}

\vspace{-2mm}
\label{table:seqlen}
\end{table}

\paragraph{Visualizing Sequence Length}Figure~\ref{fig:model_seq_lenth} illustrates the progressive improvement in model performance as the sequence length of ARFlow increases. The training loss curves demonstrate a clear trend: longer sequence lengths consistently result in lower training loss. Starting with a sequence length of 1, which has a similar loss to the original SiT model, each increase in sequence length leads to a noticeable reduction in training loss, with the sequence length of 10 achieving the lowest loss.

We also visualize samples from the ARFlow trained with different sequence lengths in Figure~\ref{fig:smaples_seq_lenth}. The results indicate that the model's performance improves as the sequence length increases.

\paragraph{CFG Scale}
\begin{wraptable}{r}{0.46\textwidth}
\vspace{-5mm}
\centering
\caption{Analysis of classifier-free guidance (CFG) scale on ARFlow-B/2 performance.}
\footnotesize
\resizebox{\linewidth}{!}{%
\begin{tabular}{cccccc}
\toprule
CFG & IS$\uparrow$ & FID$\downarrow$ & sFID & Prec & Rec \\
\midrule
1.0 & 36.73  & 25.46  & 7.13 & 0.5185 & 0.6489 \\
1.5 & 84.18  & 10.16  & 6.77 & 0.6607 & 0.5908 \\
2.0 & 143.36 & 5.33   & 7.42 & 0.7596 & 0.5154 \\
2.5 & 195.91 & 5.47   & 8.76 & 0.8197 & 0.4490 \\
3.0 & 237.44 & 7.26   & 10.30 & 0.8542 & 0.3917 \\
3.5 & 269.50 & 9.43   & 11.84 & 0.8761 & 0.3354 \\
\bottomrule
\end{tabular}
}
\vspace{-2mm}
\label{tab:cfg}
\end{wraptable}

The classifier-free guidance strength reveals interesting trade-offs in generation behavior. The FID improves substantially from CFG=1.0 (25.46) to CFG=2.0 (5.33). However, further increasing CFG leads to a quality-diversity trade-off: while precision continues to improve (from 0.7596 at CFG=2.0 to 0.8761 at CFG=3.5), recall decreases significantly (from 0.5154 to 0.3354). This suggests that stronger guidance produces higher quality samples but at the cost of reduced diversity.

\paragraph{Inference  Steps}
\begin{wraptable}{r}{0.46\textwidth}
\vspace{-4mm}
\centering
\footnotesize
\caption{Effect of inference steps on ARFlow-B/2 model performance.}
\label{tab:inference_steps}
\resizebox{\linewidth}{!}{
\begin{tabular}{cccccc}
\toprule
Steps & IS$\uparrow$ & FID$\downarrow$ & sFID$\downarrow$ & Prec$\uparrow$ & Rec$\uparrow$ \\
\midrule
10   & 14.16 & 80.63 & 48.21 & 0.3594 & 0.4680 \\
50   & 33.49 & 30.06 & 8.61  & 0.4966 & 0.6492 \\
100  & 35.93 & 26.84 & 7.54  & 0.5093 & 0.6547 \\
150  & 36.53 & 26.11 & 7.35  & 0.5121 & 0.6495 \\
250  & 36.73 & 25.46 & 7.13  & 0.5185 & 0.6489 \\
500  & 37.48 & 25.19 & 7.16  & 0.5176 & 0.6597 \\
\bottomrule
\end{tabular}
}
\vspace{-4mm}
\end{wraptable}

The number of inference steps significantly impacts generation quality. With only 10 steps, performance is poor (FID = 80.63), but increasing to 50 steps dramatically improves the FID to 30.06. Further improvements are seen up to 250 steps (FID = 25.46), after which the gains diminish, as 500 steps achieve only a marginally better FID of 25.19. Thus, we select 250 steps as the default configuration, balancing quality and inference speed.

\paragraph{Remove Cache State}
\begin{wraptable}{r}{0.46\textwidth}
\vspace{-4mm}
\centering
\footnotesize
\caption{Ablation study comparing ARFlow-B/2 model performance with and without cached states.}
\label{tab:Cache}
\resizebox{\linewidth}{!}{
\begin{tabular}{cccccc}
\toprule
Cache & Cfg & IS$\uparrow$ & FID$\downarrow$ & Prec$\uparrow$ & Rec$\uparrow$ \\
\midrule
w  & 1.0 & 36.73 & 25.46 & 0.5185 & 0.6489 \\
wo & 1.0 & 13.69 & 65.33 & 0.2517 & 0.4650 \\
\midrule
w  & 1.5 & 84.18 & 10.16 & 0.6607 & 0.5908 \\
wo & 1.5 & 18.53 & 49.76 & 0.3671 & 0.4581 \\
\bottomrule
\end{tabular}
}
\vspace{-3mm}
\end{wraptable}

Table~\ref{tab:Cache} presents an ablation study that examines the importance of cached states in our hybrid linear attention. We compare our full model with a variant where the hidden state transfer between chunks is removed, effectively eliminating the temporal memory mechanism. The results clearly demonstrate the crucial role of cached states in maintaining generation quality.

At CFG=1.0, removing the cached states leads to a dramatic degradation in performance across all metrics. The FID significantly increases from 25.46 to 65.33, while IS drops sharply from 36.73 to 13.69. This substantial performance gap becomes even more pronounced when applying classifier-free guidance (CFG=1.5), where the full model achieves an FID of 10.16 compared to 49.76 without cached states. The severe degradation in both Precision (0.6607 to 0.3671) and Recall (0.5908 to 0.4581) metrics indicates that the cached state mechanism is essential for both generation quality and diversity. It is also shown that our model learns the semantic information passed between steps.

\section{Conclusion}
We presented ARFlow, a novel framework that integrates autoregressive modeling with flow models. Our key innovations lie in two aspects: a novel sequence construction method that creates causally-ordered sequences from multiple images within the same semantic category, and a hybrid linear attention mechanism designed for efficient long-range modeling.  Our approach effectively addresses the long-range dependency limitations of traditional flow models. Extensive experiments on ImageNet demonstrate ARFlow's superior performance, achieving FID scores of 4.34 with classifier-free guidance 1.5 much better than SiT on 400k training steps. We believe the pipeline introduced in ARFlow could be extended to other generative tasks and potentially inspire new approaches for combining autoregressive and flow-based models.

\bibliographystyle{unsrt}
\bibliography{main}
\clearpage
\section*{Technical Appendices and Supplementary Material}
\section{License}Our experiment is entirely based on ImageNet (Custom License, as viewed on https://www.image-net.org/download).
And our code is mainly based on SiT(MIT license)~\cite{ma2024sit}.

\section{Visualizing ARFlow Scale} As shows in~\ref{fig:model_seq_lenth_loss},  we visualize the training loss of SIT and ARFlow for different scales, and we can find that for the same model size, ARFlow's loss is lower than that of the original SiT, and in particular, ARFlow-B5's loss is close to SiT-XL's loss.

\begin{figure}[htbp]
\centering
\includegraphics[width=\linewidth]{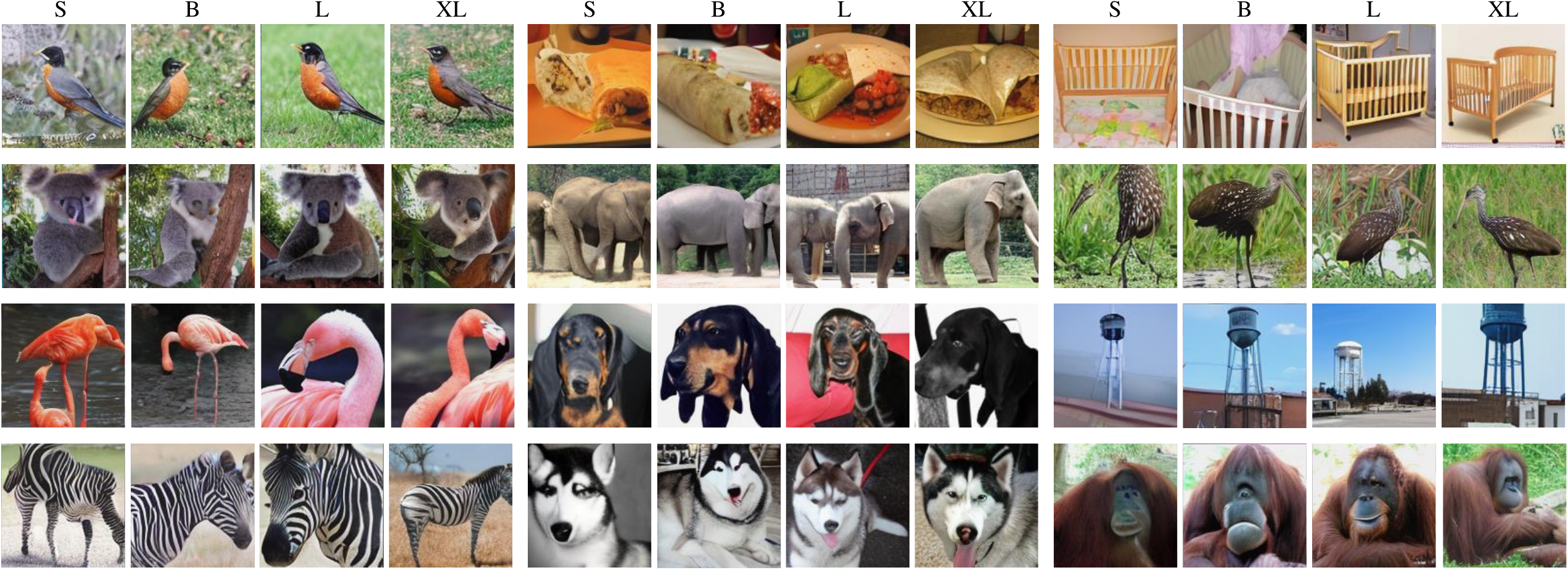}
\caption{Visual comparison of samples between different size ARFlow with CFG=4.0, demonstrating progressive improvement in sample quality as model size increase.}
\label{fig:samlpe_scale}
\vspace{3mm}
\end{figure}

Figure\ref{fig:samlpe_scale} visualizes the sampling capabilities of ARFlow at 400K training steps, across different model sizes, using the same starting noise and seed. All images are generated with CFG=4.0. The comparison shows a clear improvement in image quality and fidelity as the model size increases.

\section{Complexity Analysis of Attention Mechanisms}

\subsection{Standard Softmax Attention}

The standard softmax attention mechanism operates on an input sequence $\mathbf{X} \in \mathbb{R}^{T \times d}$ through the following steps:
\begin{align}
\begin{split}
\mathbf{Q},\mathbf{K},\mathbf{V} &= \mathbf{X}\mathbf{W}_{Q},\mathbf{X}\mathbf{W}_{K},\mathbf{X}\mathbf{W}_{V} \\
\mathbf{O} &= \mathrm{softmax}\big((\mathbf{Q}\mathbf{K}^{\top})\odot\mathbf{M}\big)\mathbf{V}
\end{split}
\end{align}

The computational complexity consists of the following components: Computing $\mathbf{Q}$, $\mathbf{K}$, $\mathbf{V}$ requires $O(T d^2)$ operations for linear projections; computing $\mathbf{Q}\mathbf{K}^{\top}$ requires $O(T^2 d)$ operations for the attention matrix; computing softmax and final output requires $O(T^2 d)$ operations. The total complexity for standard softmax attention is $O(T^2 d + T d^2)$.

\subsection{Hybrid Linear Attention}

Our hybrid linear attention combines efficient inter-chunk processing with localized intra-chunk attention. Let $C$ be the chunk size and $N = T/C$ be the number of chunks.

\subsubsection{Inter-chunk Attention}

For each chunk $i$, we compute:
\begin{align}
\mathbf{S}_{[i+1]} &= \gamma_{i+1} \mathbf{S}_{[i]} + \mathbf{K}_{[i+1]}^\top \mathbf{V}_{[i+1]} \\
\mathbf{O}_{[i+1]}^{\text{inter}} &= \mathbf{Q}_{[i+1]} \mathbf{S}_{[i]}
\end{align}

The complexity for inter-chunk attention includes hidden state update requiring $O(C d^2)$ per chunk and inter-chunk output computation requiring $O(C d^2)$ per chunk, resulting in a total inter-chunk cost of $O(N C d^2) = O(T d^2)$.

\subsubsection{Intra-chunk Attention}

Within each chunk, we compute full attention:
\begin{equation}
\mathbf{O}_{[i+1]}^{\text{intra}} = \operatorname{softmax}\left( \mathbf{Q}_{[i+1]} \mathbf{K}_{[i+1]}^\top \right) \mathbf{V}_{[i+1]}
\end{equation}

The complexity per chunk involves attention scores computation requiring $O(C^2 d)$, softmax operation requiring $O(C^2)$, and output computation requiring $O(C^2 d)$, leading to a total intra-chunk cost of $O(C^2 d)$.

\subsubsection{Total Complexity Analysis}

Combining all components:
\begin{align}
\text{Total Cost} &= N \times (O(C d^2) + O(C^2 d)) \\
&= O(T d^2) + O(T C d)
\end{align}

Since $C$ is constant:
\begin{equation}
\text{Total Cost} = O(T d^2)
\end{equation}

\subsection{Memory Complexity}

The memory requirements consist of $O(d^2)$ for the inter-chunk hidden state and $O(C^2)$ per chunk for attention scores, resulting in a total memory complexity of $O(T d) + O(T C) = O(T d)$.

\subsection{Comparison}

\begin{table}[h]
\centering
\begin{tabular}{lcc}
\toprule
\textbf{Method} & \textbf{Computational} & \textbf{Memory} \\
\midrule
Standard Softmax & $O(T^2 d + T d^2)$ & $O(T^2)$ \\
Hybrid Linear & $O(T d^2)$ & $O(T d)$ \\
\bottomrule
\end{tabular}
\caption{Complexity comparison between attention mechanisms}
\end{table}

The hybrid linear attention achieves linear complexity in sequence length $T$ while maintaining the ability to model both local and global dependencies through its dual attention mechanism. This represents a significant improvement over the quadratic complexity of standard softmax attention, making it more suitable for processing long sequences while preserving the essential ability to capture both short-range and long-range dependencies in the input data.

\section{Additional Implementation Details}
We use the same model architecture as the original SiT~\cite{ma2024sit}, but replace the standard multi-head attention mechanism with our proposed Hybrid Linear Attention. This introduces two additional linear layers each attention layer, resulting in a slight increase(5\%) in the number of parameters compared to the original model, as detailed in~\ref{tab:ARFlow}.

\begin{table}[h!]
\centering
\begin{tabular}{lccccc}
\toprule
Model & Layers & Hidden size & Heads & Parameters \\
\midrule
ARFlow-S/2  & 12 & 384  &  6  & 34M \\
ARFlow-B/2  & 12 & 768  & 12  & 137M \\
ARFlow-L/2  & 24 & 1024 & 16  & 483M \\
ARFlow-XL/2 & 28 & 1152 & 16  & 712M \\
\bottomrule
\end{tabular}%
\caption{Model size of different ARFlow models.}
\label{tab:ARFlow}
\end{table}

\section{Model Samples}

\begin{figure*}[htbp]
\centering
\includegraphics[width=0.85\linewidth]{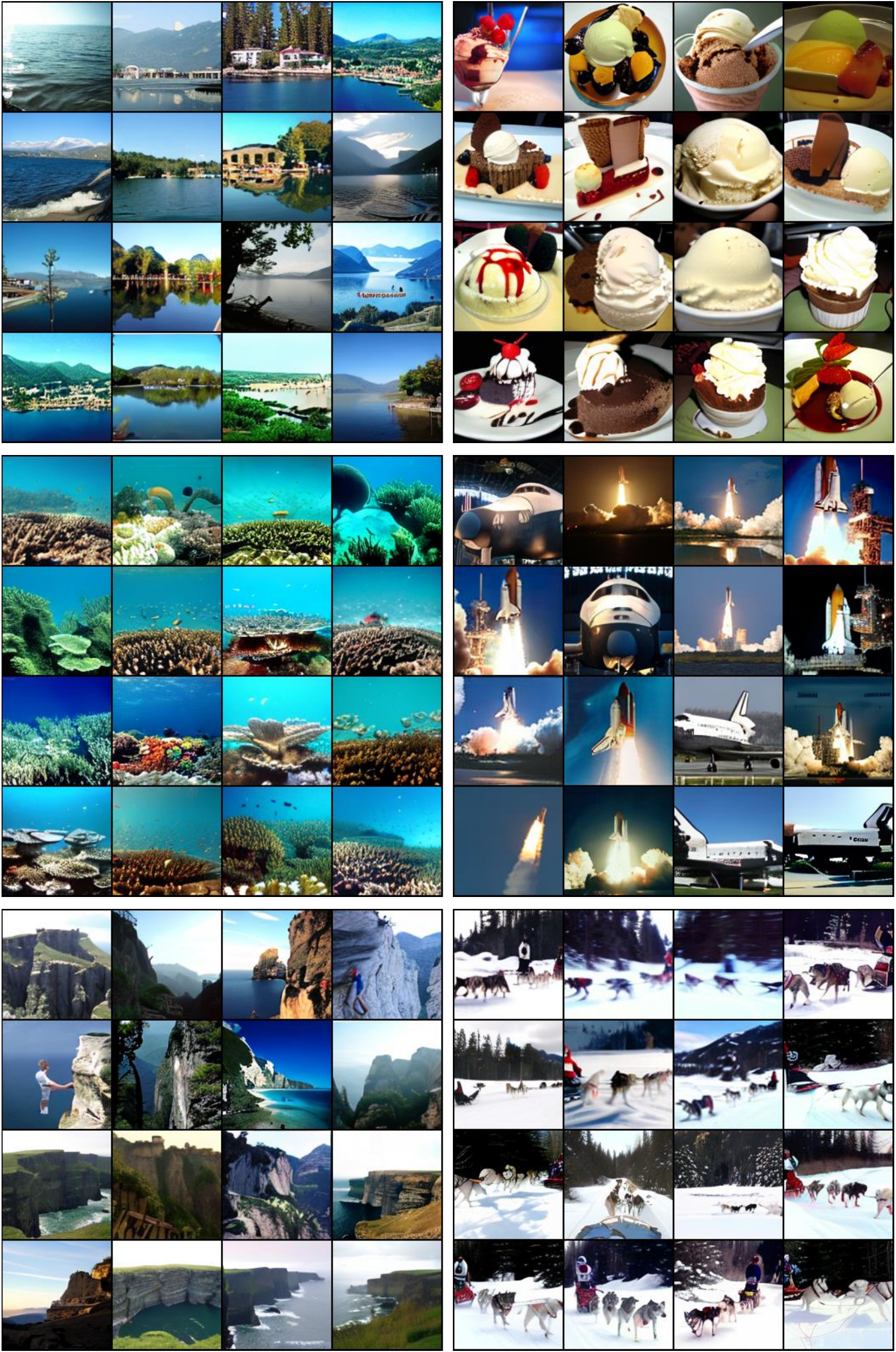}
\caption{Uncurated 128 $\times$ 128  ARFlow-XL/2 samples.}
\label{fig:1}
\end{figure*}

\begin{figure*}[htbp]
\centering
\includegraphics[width=0.85\linewidth]{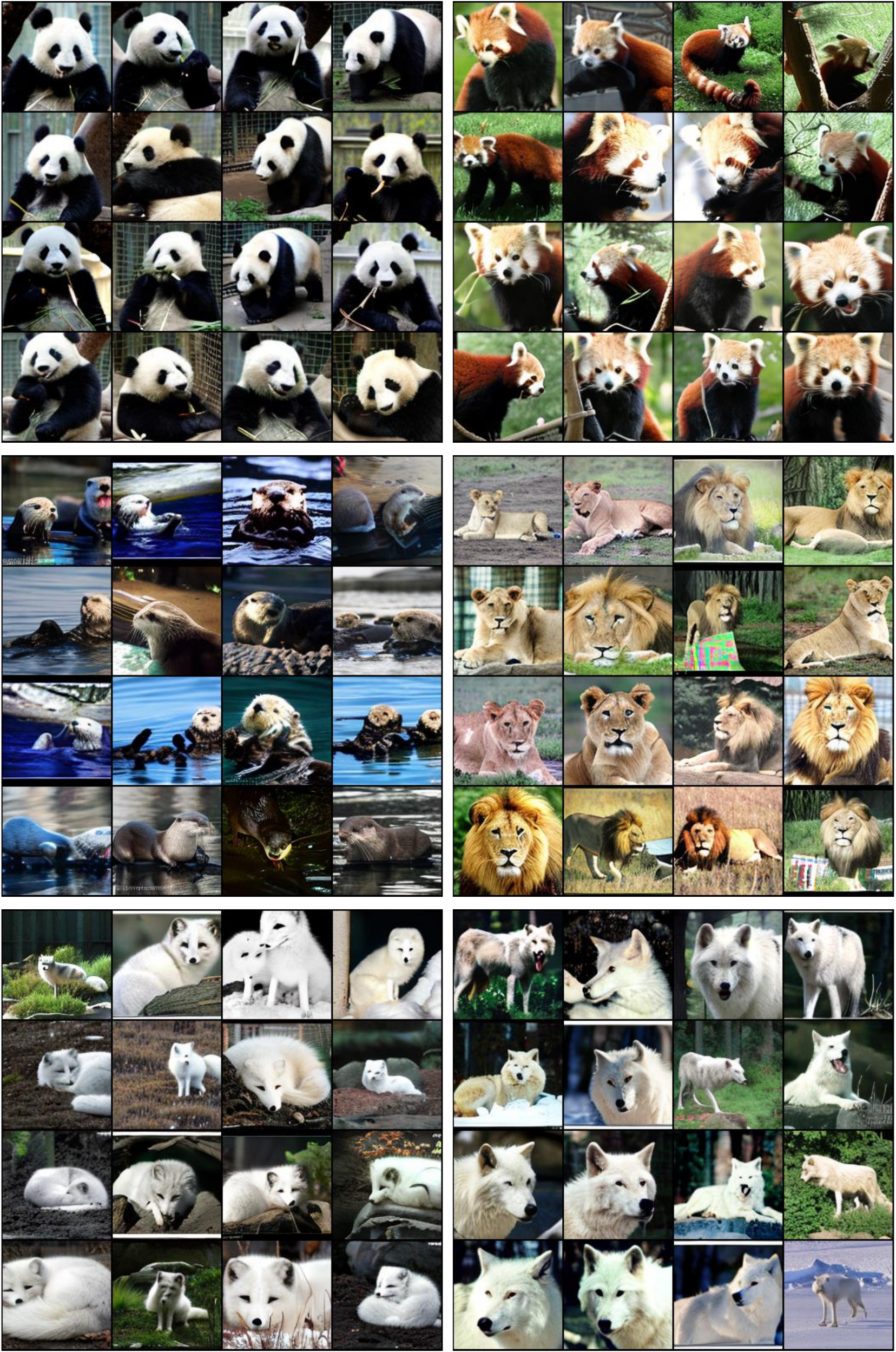}
\caption{Uncurated 128 $\times$ 128  ARFlow-XL/2 samples.}
\label{fig:2}
\end{figure*}

\begin{figure*}[htbp]
\centering
\includegraphics[width=0.85\linewidth]{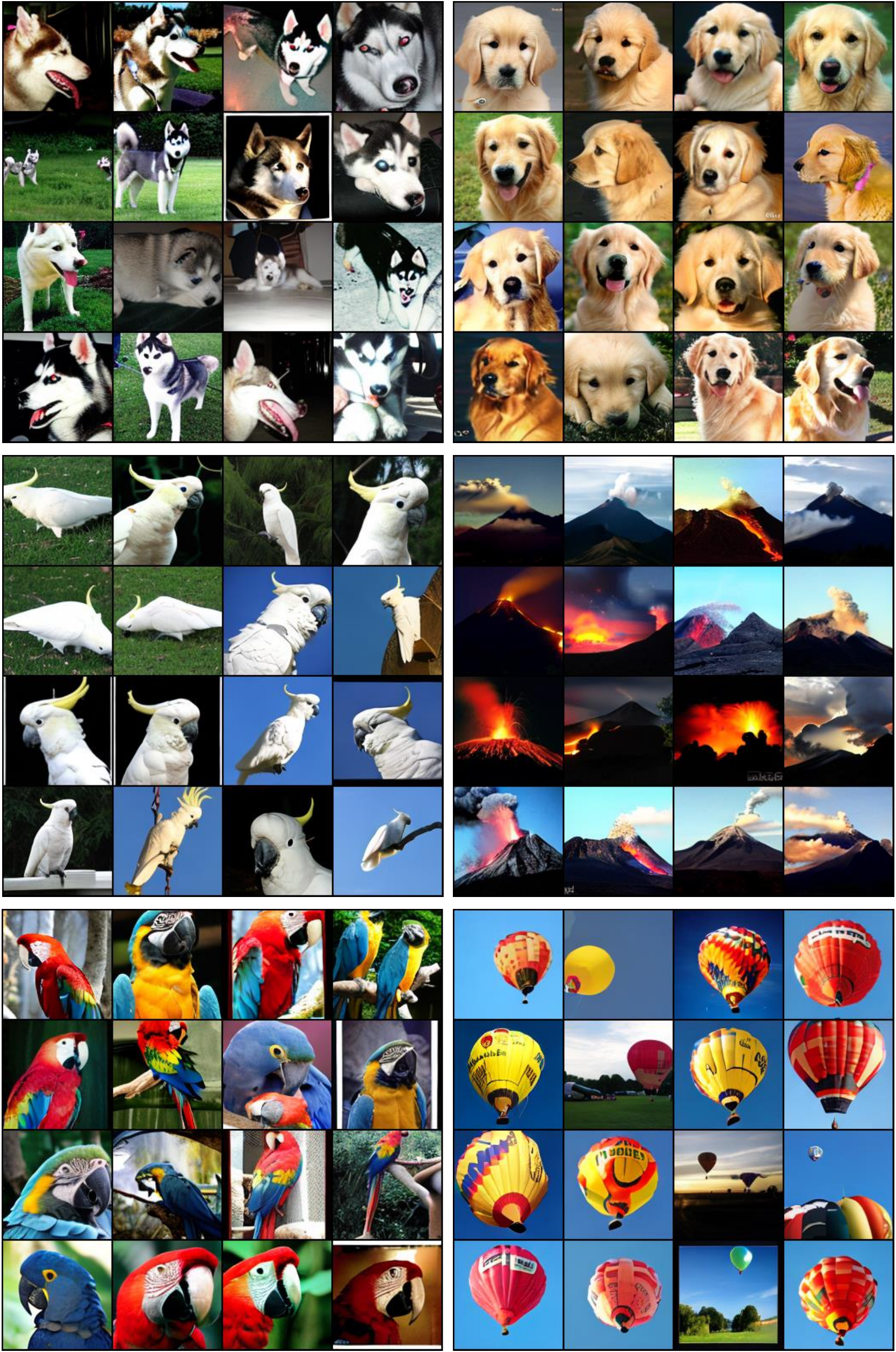}
\caption{Uncurated 128 $\times$ 128  ARFlow-XL/2 samples.}
\label{fig:3}
\end{figure*}

We show uncurated samples from our ARFlow-XL/2 model trained for 800k steps in Figure~\ref{fig:1},~\ref{fig:2},and~\ref{fig:3}, all sampled with 250 steps and  4.0 classifier free guidance scale.
\end{document}